\documentclass[twoside,english,compsoc]{IEEEtran}
\usepackage[T1]{fontenc}
\usepackage[latin1]{inputenc}
\usepackage{color}
\usepackage{array}
\usepackage{units}
\usepackage{multirow}
\usepackage{amsmath}
\usepackage{amssymb}
\usepackage{graphicx}
\usepackage{esint}

\makeatletter

\@ifundefined{showcaptionsetup}{}{%
 \PassOptionsToPackage{caption=false}{subfig}}
\usepackage{subfig}
\makeatother

\usepackage{babel}
\begin{document}

\title{The Kernel Pitman-Yor Process}

\author{Sotirios P. Chatzis, Dimitrios Korkinof, and Yiannis Demiris%
\thanks{S.P.C. and D.K. have equal contributions to this work.%
}}
\maketitle
\begin{abstract}
In this work, we propose the kernel Pitman-Yor process (KPYP) for
nonparametric clustering of data with general spatial or temporal
interdependencies. The KPYP is constructed by first introducing an
infinite sequence of random locations. Then, based on the stick-breaking
construction of the Pitman-Yor process, we define a predictor-dependent
random probability measure by considering that the discount hyperparameters
of the Beta-distributed random weights (stick variables) of the process
are not uniform among the weights, but controlled by a kernel function
expressing the proximity between the location assigned to each weight
and the given predictors. \end{abstract}

\section{Introduction}

Nonparametric Bayesian modeling techniques, especially Dirichlet process
mixture (DPM) models, have become very popular in statistics over
the last few years, for performing nonparametric density estimation
\cite{dpm2,dpm3,dpm4}. This theory is based on the observation that
an infinite number of component distributions in an ordinary finite
mixture model (clustering model) tends on the limit to a Dirichlet
process (DP) prior \cite{dpm3,DPM}. Eventually, the nonparametric
Bayesian inference scheme induced by a DPM model yields a posterior
distribution on the proper number of model component densities (inferred
clusters) \cite{bj}, rather than selecting a fixed number of mixture
components. Hence, the obtained nonparametric Bayesian formulation
eliminates the need of doing inference (or making arbitrary choices)
on the number of mixture components (clusters) necessary to represent
the modeled data.

An interesting alternative to the Dirichlet process prior for nonparametric
Bayesian modeling is the Pitman-Yor process (PYP) prior \cite{py}.
Pitman-Yor processes produce power-law distributions that allow for
better modeling populations comprising a high number of clusters with
low popularity and a low number of clusters with high popularity \cite{goldwater}.
Indeed, the Pitman-Yor process prior can be viewed as a generalization
of the Dirichlet process prior, and reduces to it for a specific selection
of its parameter values. In \cite{sudderth}, a Gaussian process-based
coupled PYP method for joint segmentation of multiple images is proposed. 

A different perspective to the problem of nonparametric data modeling
was introduced in \cite{ksbp}, where the authors proposed the kernel
stick-breaking process (KSBP). The KSBP imposes the assumption that
clustering is more probable if two feature vectors are close in a
prescribed (general) space, which may be associated explicitly with
the spatial or temporal position of the modeled data. This way, the
KSBP is capable of exploiting available prior information regarding
the spatial or temporal relations and dependencies between the modeled
data. 

Inspired by these advances, and motivated by the interesting properties
of the PYP, in this paper we come up with a different approach towards
predictor-dependent random probability measures for non-parametric
Bayesian clustering. We first introduce an infinite sequence of random
spatial or temporal locations. Then, based on the stick-breaking construction
of the Pitman-Yor process, we define a predictor-dependent random
probability measure by considering that the discount hyperparameters
of the Beta-distributed random weights (stick variables) of the process
are not uniform among the weights, but controlled by a kernel function
expressing the proximity between the location assigned to each weight
and the given predictors. The obtained random probability measure
is dubbed the kernel Pitman-Yor process (KPYP) for non-parametric
clustering of data with general spatial or temporal interdependencies.
We empirically study the performance of the KPYP prior in unsupervised
image segmentation and text-dependent speaker identification, and
compare it to the kernel stick-breaking process, and the Dirichlet
process prior. 

The remainder of this paper is organized as follows: In Section 2,
we provide a brief presentation the Pitman-Yor process, as well as
the kernel stick-breaking process, and its desirable properties in
clustering data with spatial or temporal dependencies. In Section
3, the proposed nonparametric prior for clustering data with temporal
or spatial dependencies is introduced, its relations to existing methods
are discussed, and an efficient variational Bayesian algorithm for
model inference is derived.

\section{Theoretical Background}

\subsection{The Pitman-Yor Process}

Dirichlet process (DP) models were first introduced by Ferguson \cite{ferguson-dp}.
A DP is characterized by a base distribution $G_{0}$ and a positive
scalar $\alpha$, usually referred to as the innovation parameter,
and is denoted as $\mathrm{DP}(\alpha,G_{0})$. Essentially, a DP
is a distribution placed over a distribution. Let us suppose we randomly
draw a sample distribution $G$ from a DP, and, subsequently, we independently
draw $M$ random variables $\{\Theta_{m}^{*}\}_{m=1}^{M}$ from $G$:
\begin{equation}
G|\alpha,G_{0}\sim\mathrm{DP}(\alpha,G_{0})
\end{equation}
\begin{equation}
\Theta_{m}^{*}|G\sim G,\quad m=1,\dots M
\end{equation}
Integrating out $G$, the joint distribution of the variables $\{\Theta_{m}^{*}\}_{m=1}^{M}$
can be shown to exhibit a clustering effect. Specifically, given the
first $M-1$ samples of $G$, $\{\Theta_{m}^{*}\}_{m=1}^{M-1}$, it
can be shown that a new sample $\Theta_{M}^{*}$ is either (a) drawn
from the base distribution $G_{0}$ with probability $\frac{\alpha}{\alpha+M-1}$,
or (b) is selected from the existing draws, according to a multinomial
allocation, with probabilities proportional to the number of the previous
draws with the same allocation \cite{polya-urn}. Let $\{\Theta_{c}\}_{c=1}^{C}$
be the set of distinct values taken by the variables $\{\Theta_{m}^{*}\}_{m=1}^{M-1}$.
Denoting as $f_{c}^{M-1}$ the number of values in $\{\Theta_{m}^{*}\}_{m=1}^{M-1}$
that equal to $\Theta_{c}$, the distribution of $\Theta_{M}^{*}$
given $\{\Theta_{m}^{*}\}_{m=1}^{M-1}$ can be shown to be of the
form \cite{polya-urn}
\begin{equation}
\begin{aligned}p(\Theta_{M}^{*}|\{\Theta_{m}^{*}\}_{m=1}^{M-1},\alpha,G_{0})= & \frac{\alpha}{\alpha+M-1}G_{0}\\
 & +\sum_{c=1}^{C}\frac{f_{c}^{M-1}}{\alpha+M-1}\delta_{\Theta_{c}}
\end{aligned}
\end{equation}
where $\delta_{\Theta_{c}}$ denotes the distribution concentrated
at a single point $\Theta_{c}$. 

The Pitman-Yor process \cite{py} functions similar to the Dirichlet
process. Let us suppose we randomly draw a sample distribution $G$
from a PYP, and, subsequently, we independently draw $M$ random variables
$\{\Theta_{m}^{*}\}_{m=1}^{M}$ from $G$:
\begin{equation}
G|d,\alpha,G_{0}\sim\mathrm{PY}(d,\alpha,G_{0})
\end{equation}
 with
\begin{equation}
\Theta_{m}^{*}|G\sim G,\quad m=1,\dots M\;
\end{equation}
where $d\in[0,1)$ is the discount parameter of the Pitman-Yor process,
$\alpha>-d$ is its innovation parameter, and $G_{0}$ the base distribution.
Integrating out $G$, similar to Eq. (3), we now yield
\begin{equation}
\begin{aligned}p(\Theta_{M}^{*}|\{\Theta_{m}^{*}\}_{m=1}^{M-1},d,\alpha,G_{0})= & \frac{\alpha+dC}{\alpha+M-1}G_{0}\\
 & +\sum_{c=1}^{C}\frac{f_{c}^{M-1}-d}{\alpha+M-1}\delta_{\Theta_{c}}
\end{aligned}
\end{equation}

As we observe, the PYP yields an expression for $p(\Theta_{M}^{*}|\{\Theta_{m}^{*}\}_{m=1}^{M-1},G_{0})$
quite similar to that of the DP, also possessing the rich-gets-richer
clustering property, i.e., the more samples have been assigned to
a draw from $G_{0}$, the more likely subsequent samples will be assigned
to the same draw. Further, the more we draw from $G_{0}$, the more
likely a new sample will again be assigned to a new draw from $G_{0}$.
These two effects together produce a \emph{power-law distribution}
where many unique $\Theta_{m}^{*}$ values are observed, most of them
rarely \cite{py}. In particular, for $d>0$, the number of unique
values scales as $\mathcal{O}(\alpha M^{d})$, where $M$ is the total
number of draws. Note also that, for $d=0$, the Pitman-Yor process
reduces to the Dirichlet process, in which case the number of unique
values grows more slowly at $\mathcal{O}(\alpha\mathrm{log}M)$ \cite{teh}.

A characterization of the (unconditional) distribution of the random
variable $G$ drawn from a PYP, $\mathrm{PY}(d,\alpha,G_{0})$, is
provided by the stick-breaking construction of Sethuraman \cite{stick-break}.
Consider two infinite collections of independent random variables
$\boldsymbol{v}={(v_{c})}_{c=1}^{\infty}$, $\{\Theta_{c}\}_{c=1}^{\infty}$,
where the $v_{c}$ are drawn from a Beta distribution, and the $\Theta_{c}$
are independently drawn from the base distribution $G_{0}$. The stick-breaking
representation of $G$ is then given by \cite{teh} 
\begin{equation}
G=\sum_{c=1}^{\infty}\varpi_{c}(\boldsymbol{v})\delta_{\Theta_{c}}
\end{equation}
 where
\begin{equation}
p(v_{c})=\mathrm{Beta}(v_{c}|1-d,\alpha+dc)
\end{equation}
\begin{equation}
\boldsymbol{v}=\left(v_{c}\right)_{c=1}^{\infty}
\end{equation}
\begin{equation}
\varpi_{c}(\boldsymbol{v})=v_{c}\prod_{j=1}^{c-1}(1-v_{j})\quad\in[0,1]
\end{equation}
 and
\begin{equation}
\sum_{c=1}^{\infty}\varpi_{c}(\boldsymbol{v})=1
\end{equation}

\subsection{The Kernel Stick-Breaking Process}

An alternative to the above approaches, allowing for taking into account
additional prior information regarding spatial or temporal dependencies
in the modeled datasets, is the kernel stick-breaking process introduced
in \cite{ksbp}. The basic notion in the formulation of the KSBP consists
in the introduction of a predictor-dependent prior, which promotes
clustering of adjacent data points in a prescribed (general) space. 

Let us consider that the observed data points $\boldsymbol{y}\in\mathcal{Y}$
are associated with positions where measurement was taken $\boldsymbol{x}\in\mathcal{X}$,
arranged on a $D$-dimensional lattice. For example, in cases of sequential
data modeling, the observed data points $\boldsymbol{y}$ are naturally
associated with an one-dimensional lattice that depicts their temporal
succession, i.e. the time point these measurements were taken. In
cases of computer vision applications, we might be dealing with observations
$\boldsymbol{y}$ measured on different locations on a two-dimensional
or three-dimensional space $\mathcal{X}$. 

To take this prior information into account, the KSBP postulates that
the random process $G$ in (1) comprises a function of the predictors
$\boldsymbol{x}$ related to the observable data points $\boldsymbol{y}$,
expressing their location in the prescribed space $\mathcal{X}$.
Specifically, it is assumed that
\begin{equation}
G=\sum_{c=1}^{\infty}\varpi_{c}(\boldsymbol{v}(\boldsymbol{x}))\delta_{\Theta_{c}}
\end{equation}
 where
\begin{equation}
\varpi_{c}(\boldsymbol{v}(\boldsymbol{x}))=v_{c}(\boldsymbol{x},\Gamma_{c};\psi_{c})\prod_{j=1}^{c-1}(1-v_{j}(\boldsymbol{x},\Gamma_{j};\psi_{j}))\quad\in[0,1]
\end{equation}
\begin{equation}
\boldsymbol{v}(\boldsymbol{x})=\left(v_{c}(\boldsymbol{x},\Gamma_{c};\psi_{c})\right)_{c=1}^{\infty}
\end{equation}
\begin{equation}
v_{c}(\boldsymbol{x},\Gamma_{c};\psi_{c})=V_{c}k(\boldsymbol{x},\Gamma_{c};\psi_{c})
\end{equation}
\begin{equation}
p(V_{c})=\mathrm{Beta}(V_{c}|1,\alpha)
\end{equation}
and $k(\boldsymbol{x},\Gamma_{c};\psi_{c})$ is a kernel function
centered at $\Gamma_{c}$ with hyperparameter $\psi_{c}$. 

By selecting an appropriate form of the kernel function $k(\boldsymbol{x},\Gamma_{c};\psi_{c})$,
KSBP allows for obtaining prior probabilities $\varpi_{c}(\boldsymbol{v}(\boldsymbol{x}))$
for the derived clusters that depend on the values of the predictors
(spatial or temporal locations) $\boldsymbol{x}$. Indeed, the closer
the location $\boldsymbol{x}$ of an observation $\boldsymbol{y}$
is to the location $\Gamma_{c}$ assigned to the $c$th cluster, the
higher the prior probability $\varpi_{c}(\boldsymbol{v}(\boldsymbol{x}))$
becomes. Thus, the KSBP prior promotes by construction clustering
of (spatially or temporally) adjacent data points. For example, a
typical selection for the kernel $k(\boldsymbol{x},\Gamma_{c};\psi_{c})$
is the radial basis function (RBF) kernel
\begin{equation}
k(\boldsymbol{x},\Gamma_{c};\psi_{c})=\mathrm{exp}\left[-\frac{||\boldsymbol{x}-\Gamma_{c}||^{2}}{\psi_{c}^{2}}\right]
\end{equation}

\section{Proposed Approach}

\subsection{Model Formulation}

We aim to obtain a clustering algorithm which takes into account the
prior information regarding the (temporal or spatial) adjacencies
of the observed data in the locations space $\mathcal{X}$, promoting
clustering of data adjacent in the space $\mathcal{X}$, and discouraging
clustering of data points relatively near in the feature space $\mathcal{Y}$
but far in the locations space $\mathcal{X}$. For this purpose, we
seek to provide a location-dependent nonparametric prior for clustering
the observed data $\boldsymbol{y}$.

Motivated by the definition and the properties of the Pitman-Yor process
discussed in the previous section, to effect these goals, in this
work we introduce a random probability measure $G(\boldsymbol{x})$
under which, given the first $M-1$ samples $\{\Theta_{m}^{*}\}_{m=1}^{M-1}$
drawn from $G$, a new sample $\Theta_{M}^{*}$ associated with a
measurement location $\boldsymbol{x}$ is distributed according to
\begin{equation}
\begin{aligned}p(\Theta_{M}^{*}| & \{\Theta_{m}^{*}\}_{m=1}^{M-1};\boldsymbol{x},k,\alpha,\hat{X},G_{0})\\
= & \frac{\alpha+\sum_{c=1}^{C}\left[1-k(\boldsymbol{x},\hat{\boldsymbol{x}}_{c};\psi_{c})\right]}{\alpha+M-1}G_{0}\\
 & +\sum_{c=1}^{C}\frac{f_{c}^{M-1}+k(\boldsymbol{x},\hat{\boldsymbol{x}}_{c};\psi_{c})-1}{\alpha+M-1}\delta_{\Theta_{c}}
\end{aligned}
\end{equation}
where $f_{c}^{M-1}$ is the number of values in $\{\Theta_{m}^{*}\}_{m=1}^{M-1}$
that equal to $\Theta_{c}$, $\{\Theta_{c}\}_{c=1}^{C}$ is the set
of distinct values taken by the variables $\{\Theta_{m}^{*}\}_{m=1}^{M-1}$,
$G_{0}$ is the employed base measure, $\hat{\boldsymbol{x}}_{c}$
is the location assigned to the $c$th cluster, $\hat{X}\triangleq\{\hat{\boldsymbol{x}}_{c}\}_{c}$,
$k(\cdot,\hat{\boldsymbol{x}};\psi)$ \emph{is a bounded kernel function
taking values in the interval $[0,1]$, such that
\begin{equation}
\underset{\boldsymbol{x}\rightarrow\hat{\boldsymbol{x}}}{\mathrm{lim}}k(\boldsymbol{x},\hat{\boldsymbol{x}};\psi)=1
\end{equation}
\begin{equation}
\underset{\mathrm{dist}(\boldsymbol{x},\hat{\boldsymbol{x}})\rightarrow\infty}{\mathrm{lim}}k(\boldsymbol{x},\hat{\boldsymbol{x}};\psi)=0
\end{equation}
}$\alpha$ is the innovation parameter of the process, conditioned
to satisfy $\alpha>0$, and $\mathrm{dist}(\cdot,\cdot)$ is the distance
metric used by the employed kernel function. We dub this random probability
measure $G(\boldsymbol{x})$ the kernel Pitman-Yor process, and we
denote
\begin{equation}
\Theta_{m}^{*}|\boldsymbol{x};G\sim G(\boldsymbol{x}),\quad m=1,\dots M\;
\end{equation}
 with
\begin{equation}
G(\boldsymbol{x})|k,\alpha,\hat{X},G_{0}\sim\mathrm{KPYP}(\boldsymbol{x};k,\alpha,\hat{X},G_{0})
\end{equation}

The stick-breaking construction of the KPYP $G(\boldsymbol{x})$ follows
directly from the above definition (18), and the relevant discussions
of section 2. Considering a KPYP $G$ with cluster locations set $\hat{X}=\{\hat{\boldsymbol{x}}_{c}\}_{c=1}^{\infty}$,\emph{
}kernel function $k(\cdot,\cdot)$ satisfying the constraints (19)
and (20), and innovation parameter $\alpha$, we have
\begin{equation}
G(\boldsymbol{x})=\sum_{c=1}^{\infty}\varpi_{c}(\boldsymbol{v}(\boldsymbol{x}))\delta_{\Theta_{c}}
\end{equation}
 where
\begin{equation}
v_{c}(\boldsymbol{x})\sim\mathrm{Beta}\left(k(\boldsymbol{x},\hat{\boldsymbol{x}}_{c};\psi_{c}),\alpha+c\left[1-k(\boldsymbol{x},\hat{\boldsymbol{x}}_{c};\psi_{c})\right]\right)
\end{equation}
 and
\begin{equation}
\varpi_{c}(\boldsymbol{v}(\boldsymbol{x}))=v_{c}(\boldsymbol{x})\prod_{j=1}^{c-1}(1-v_{j}(\boldsymbol{x}))\quad\in[0,1]
\end{equation}
\textbf{Proposition 1. }The stochastic process\textbf{ $G(\boldsymbol{x})$
}defined in (23)-(25) is a valid random probability measure.\\
\emph{Proof. }We need to show that 
\begin{equation}
\sum_{c=1}^{\infty}\varpi_{c}(\boldsymbol{v}(\boldsymbol{x}))=1
\end{equation}
For this purpose, we follow an approach similar to \cite{ksbp}. From
(25), we have
\begin{equation}
1-\sum_{c=1}^{C-1}\varpi_{c}(\boldsymbol{v}(\boldsymbol{x}))=\prod_{c=1}^{C-1}\left[1-v_{c}(\boldsymbol{x})\right]
\end{equation}
 Then, in the limit as $C\rightarrow\infty$, and taking logs in both
sides of (27), we have 
\begin{equation}
\sum_{c=1}^{\infty}\varpi_{c}(\boldsymbol{v}(\boldsymbol{x}))=1\mathrm{\; if\; and\; only\; if}\;\sum_{c=1}^{\infty}\mathrm{log}\left[1-v_{c}(\boldsymbol{x})\right]=-\infty
\end{equation}
Based on Kolmogorov three series theorem, the summation on the right
is over independent random variables and is equal to $-\infty$ if
and only if $\sum_{c=1}^{\infty}\mathbb{E}\big\{\mathrm{log}\left[1-v_{c}(\boldsymbol{x})\right]\big\}=-\infty$.
However, $v_{c}(\boldsymbol{x})$ follows a Beta distribution, which
means $v_{c}(\boldsymbol{x})\in[0,1]$, thus $\mathrm{log}\left[1-v_{c}(\boldsymbol{x})\right]\leq0$,
and hence its expectation is negative; thus, the condition is satisfied,
and (26) holds true.\\

\subsection{Relation to the KSBP}

Indeed, the proposed KPYP shares some common ideas with the KSBP of
\cite{ksbp}. The KSBP considers that
\begin{equation}
G(\boldsymbol{x})=\sum_{c=1}^{\infty}\varpi_{c}(\boldsymbol{v}(\boldsymbol{x}))\delta_{\Theta_{c}}\;
\end{equation}
 where
\begin{equation}
\varpi_{c}(\boldsymbol{v}(\boldsymbol{x}))=v_{c}(\boldsymbol{x})\prod_{j=1}^{c-1}(1-v_{j}(\boldsymbol{x}))\in[0,1]
\end{equation}
\begin{equation}
v_{c}(\boldsymbol{x})=V_{c}k(\boldsymbol{x},\hat{\boldsymbol{x}}_{c};\psi_{c})
\end{equation}
\begin{equation}
p(V_{c})=\mathrm{Beta}\left(V_{c}|1,\alpha\right)
\end{equation}
From this definition, we observe that there is a key difference between
the KPYP and the KSBP: the KSBP multiplies stick variables sharing
the same Beta prior with a bounded kernel function centered at a location
$\hat{\boldsymbol{x}}$ unique for each stick, to obtain a predictor
(location)-dependent random probability measure. Instead, the KPYP
considers stick variables with different Beta priors, with the prior
of each stick variable employing a different ``discount hyperparameter,''
defined as a bounded kernel centered at a location $\hat{\boldsymbol{x}}$
unique for each stick. This way, the KPYP controls the assignment
of observations to clusters by discounting clusters the centers of
which are too far from the clustered data points in the locations
space $\mathcal{X}$. 

It is interesting to compute the mean and variance of the stick variables
$v_{c}(\boldsymbol{x})$ for these two stochastic processes, \emph{for
a given observation location $\boldsymbol{x}$ and cluster center
$\hat{\boldsymbol{x}}_{c}$.} In the case of the KPYP, we have
\begin{equation}
\mathbb{E}\big[v_{c}(\boldsymbol{x})\big]=\frac{k(\boldsymbol{x},\hat{\boldsymbol{x}}_{c};\psi_{c})}{k(\boldsymbol{x},\hat{\boldsymbol{x}}_{c};\psi_{c})+\alpha_{c}}
\end{equation}
\begin{equation}
\mathbb{V}\big[v_{c}(\boldsymbol{x})\big]=\frac{k(\boldsymbol{x},\hat{\boldsymbol{x}}_{c};\psi_{c})\alpha_{c}}{\left(k(\boldsymbol{x},\hat{\boldsymbol{x}}_{c};\psi_{c})+\alpha_{c}\right)^{2}\left(k(\boldsymbol{x},\hat{\boldsymbol{x}}_{c};\psi_{c})+\alpha_{c}+1\right)}
\end{equation}
 where
\begin{equation}
\alpha_{c}\triangleq\alpha+c\left(1-k(\boldsymbol{x},\hat{\boldsymbol{x}}_{c};\psi_{c})\right)
\end{equation}
 On the contrary, for the KSBP we have 
\begin{equation}
\mathbb{E}\big[v_{c}(\boldsymbol{x})\big]=\frac{k(\boldsymbol{x},\hat{\boldsymbol{x}}_{c};\psi_{c})}{1+\alpha}
\end{equation}
\begin{equation}
\mathbb{V}\big[v_{c}(\boldsymbol{x})\big]=\frac{k(\boldsymbol{x},\hat{\boldsymbol{x}}_{c};\psi_{c})^{2}\alpha}{\left(1+\alpha\right)^{2}\left(\alpha+2\right)}
\end{equation}
From (33) and (36), we observe that the for a \emph{given} \emph{observation
location $\boldsymbol{x}$ and cluster center $\hat{\boldsymbol{x}}_{c}$,}
same increase in the value of the kernel function $k(\boldsymbol{x},\hat{\boldsymbol{x}}_{c};\psi_{c})$
induces a much greater increase in the expected value of the stick
variable $v_{c}(\boldsymbol{x})$ employed by the KPYP compared to
the increase in the expectation of the stick variable $v_{c}(\boldsymbol{x})$
employed by the KSBP. Hence, the predictor (location)-dependent prior
probabilities of cluster assignment of the KPYP appear to vary more
steeply with the employed kernel function values compared to the KSBP.

\subsection{Variational Bayesian Inference}

Inference for nonparametric models can be conducted under a Bayesian
setting, typically by means of variational Bayes (e.g., \cite{vbdpm}),
or Monte Carlo techniques (e.g., \cite{dpm-hmm}). Here, we prefer
a variational Bayesian approach, due to its better computational costs.
For this purpose, we additionally impose a Gamma prior over the innovation
parameter $\alpha$, with
\begin{equation}
p(\alpha)=\mathcal{G}(\alpha|\eta_{1},\eta_{2}).
\end{equation}

Let us a consider a set of observations $Y=\{\boldsymbol{y}_{n}\}_{n=1}^{N}$
with corresponding locations $X=\{\boldsymbol{x}_{n}\}_{n=1}^{N}$.
We postulate for our observed data a likelihood function of the form
\begin{equation}
p(\boldsymbol{y}_{n}|z_{n}=c)=p(\boldsymbol{y}_{n}|\boldsymbol{\theta}_{c})
\end{equation}
where the hidden variables $z_{n}$ are defined such that $z_{n}=c$
if the $n$th data point is considered to be derived from the $c$th
cluster. We impose a multinomial prior over the hidden variables $z_{n}$,
with
\begin{equation}
p(z_{n}=c|\boldsymbol{x}_{n})=\varpi_{c}(\boldsymbol{v}(\boldsymbol{x}_{n}))
\end{equation}
where the $\varpi_{c}(\boldsymbol{v}(\boldsymbol{x}))$ are given
by (25), with the prior over the $v_{c}(\boldsymbol{x})$ given by
(24). We also impose a suitable conjugate exponential prior over the
likelihood parameters $\boldsymbol{\theta}_{c}$. 

Our variational Bayesian inference formalism consists in derivation
of a family of variational posterior distributions $q(.)$ which approximate
the true posterior distribution over $\{z_{n}\}_{n=1}^{N}$, $\{\boldsymbol{v}(\boldsymbol{x}_{n})\}_{n=1}^{N}$,
and $\{\boldsymbol{\theta}_{c}\}_{c=1}^{\infty}$, and the innovation
parameter $\alpha$. Apparently, under this infinite dimensional setting,
Bayesian inference is not tractable. For this reason, we fix a value
$C$ and we let the variational posterior over the $v_{i}(\boldsymbol{x})$
have the property $q(v_{C}(\boldsymbol{x})=1)=1,\;\forall\boldsymbol{x}\in\mathcal{X}$,
i.e. we set $\varpi_{c}(\boldsymbol{v}(\boldsymbol{x}))$ equal to
zero for $c>C$, $\forall\boldsymbol{x}\in\mathcal{X}$. 

Let $W=\{\alpha,\{z_{n}\}_{n=1}^{N},\{(v_{c}(\boldsymbol{x}_{n}))_{c=1}^{C}\}_{n=1}^{N},\{\boldsymbol{\theta}_{c}\}_{c=1}^{C}\}$
be the set of the parameters of our truncated model over which a prior
distribution has been imposed, and $\Xi$ be the set of the hyperparameters
of the model, comprising the $\{\psi_{c}\}_{c=1}^{C}$ and the hyperparameters
of the priors imposed over the innovation parameter $\alpha$ and
the likelihood parameters $\boldsymbol{\theta}_{c}$ of the model.
Variational Bayesian inference consists in derivation of an approximate
posterior $q(W)$ by maximization (in an iterative fashion) of the
variational free energy
\begin{equation}
\mathcal{L}(q)=\int\mathrm{d}Wq(W)\mathrm{log}\frac{p(X,Y,W|\Xi)}{q(W)}
\end{equation}
Having considered a conjugate exponential prior configuration, the
variational posterior $q(W)$ is expected to take the same functional
form as the prior, $p(W)$ \cite{BishopBook}. The variational free
energy of our model reads 
\begin{equation}
\begin{split}\mathcal{L}(q) & =\int\mathrm{d}\alpha q(\alpha)\bigg\{\mathrm{log}\frac{p(\alpha|\eta_{1},\eta_{2})}{q(\alpha)}\\
 & +\sum_{c=1}^{C-1}\sum_{n=1}^{N}\int\mathrm{d}v_{c}(\boldsymbol{x}_{n})q(v_{c}(\boldsymbol{x}_{n}))\mathrm{log}\frac{p(v_{c}(\boldsymbol{x}_{n})|\alpha)}{q(v_{c}(\boldsymbol{x}_{n}))}\bigg\}\\
+ & \sum_{c=1}^{C}\int\mathrm{d}\boldsymbol{\theta}_{c}q(\boldsymbol{\theta}_{c})\mathrm{log}\frac{p(\boldsymbol{\theta}_{c})}{q(\boldsymbol{\theta}_{c})}+\sum_{c=1}^{C}\sum_{n=1}^{N}q(z_{n}=c)\\
 & \times\bigg\{\int\mathrm{d}\boldsymbol{v}(\boldsymbol{x}_{n})q(\boldsymbol{v}(\boldsymbol{x}_{n}))\mathrm{log}p(z_{n}=c|\boldsymbol{x}_{n})\\
 & -\mathrm{log}q(z_{n}=c)+\int\mathrm{d}\boldsymbol{\theta}_{c}q(\boldsymbol{\theta}_{c})\mathrm{log}p(\boldsymbol{y}_{n}|\boldsymbol{\theta}_{c})\bigg\}
\end{split}
\end{equation}

\subsection{Variational Posteriors}

Let us denote as $\left<.\right>$ the posterior expectation of a
quantity. We have
\begin{equation}
q(v_{c}(\boldsymbol{x}_{n}))=\mathrm{Beta}(v_{c}(\boldsymbol{x}_{n})|\tilde{\beta}_{c,n},\hat{\beta}_{c,n})
\end{equation}
 where
\begin{equation}
\tilde{\beta}_{c,n}=k(\boldsymbol{x}_{n},\hat{\boldsymbol{x}}_{c};\psi_{c})+\sum_{m:\boldsymbol{x}_{m}=\boldsymbol{x}_{n}}q(z_{m}=c)
\end{equation}
\begin{equation}
\begin{aligned}\hat{\beta}_{c,n}= & \left<\alpha\right>+c\left[1-k(\boldsymbol{x}_{n},\hat{\boldsymbol{x}}_{c};\psi_{c})\right]\\
 & +\sum_{m:\boldsymbol{x}_{m}=\boldsymbol{x}_{n}}\sum_{c'=c+1}^{C}q(z_{m}=c')
\end{aligned}
\end{equation}
 and 
\begin{equation}
q(\alpha)=\mathcal{G}(\alpha|\hat{\eta}_{1},\hat{\eta}_{2})
\end{equation}
 where
\begin{equation}
\hat{\eta}_{1}=\eta_{1}+N(C-1)
\end{equation}
\begin{equation}
\hat{\eta}_{2}=\eta_{2}-\sum_{c=1}^{C-1}\sum_{n=1}^{N}\left[\psi(\hat{\beta}_{c,n})-\psi(\tilde{\beta}_{c,n}+\hat{\beta}_{c,n})\right]
\end{equation}
 $\psi(.)$ denotes the Digamma function, and
\begin{equation}
\left<\alpha\right>=\frac{\hat{\eta}_{1}}{\hat{\eta}_{2}}
\end{equation}
 Further, the cluster assignment variables yield
\begin{equation}
q(z_{nc}=1)\propto\mathrm{exp}\left(\left<\mathrm{log}\varpi_{c}(\boldsymbol{v}(\boldsymbol{x}_{n}))\right>\right)\mathrm{exp}\left(\varphi_{nc}\right)
\end{equation}
 where
\begin{equation}
\left<\mathrm{log}\varpi_{c}(\boldsymbol{v}(\boldsymbol{x}_{n}))\right>=\sum_{c'=1}^{c-1}\left<\mathrm{log}(1-v_{c'}(\boldsymbol{x}_{n}))\right>+\left<\mathrm{log}v_{c}(\boldsymbol{x}_{n})\right>
\end{equation}
\begin{equation}
\varphi_{nc}=\left<\mathrm{log}p(\boldsymbol{y}_{n}|\boldsymbol{\theta}_{c})\right>_{q(\boldsymbol{\theta}_{c})}
\end{equation}
 and 
\begin{equation}
\left<\mathrm{log}v_{c}(\boldsymbol{x}_{n})\right>=\psi(\tilde{\beta}_{c,n})-\psi(\tilde{\beta}_{c,n}+\hat{\beta}_{c,n})
\end{equation}
\begin{equation}
\left<\mathrm{log}(1-v_{c}(\boldsymbol{x}_{n}))\right>=\psi(\hat{\beta}_{c,n})-\psi(\tilde{\beta}_{c,n}+\hat{\beta}_{c,n})
\end{equation}
 Regarding the parameters $\boldsymbol{\theta}_{c}$, we obtain
\begin{equation}
\mathrm{log}q(\boldsymbol{\theta}_{c})\propto\mathrm{log}p(\boldsymbol{\theta}_{c})+\sum_{n=1}^{N}q(z_{n}=c)\mathrm{log}p(\boldsymbol{y}_{n}|\boldsymbol{\theta}_{c})
\end{equation}
Finally, regarding the model hyperparameters $\Xi$, we obtain the
hyperparameters of the employed kernel functions $\psi_{c}$ by maximization
of the lower bound $\mathcal{L}(q)$, and we heuristically select
the values of the rest.

\subsection{Learning the cluster locations $\hat{\boldsymbol{x}}_{c}$}

Regarding determination of the locations assigned to the obtained
clusters, $\hat{\boldsymbol{x}}_{c}$, these can be obtained by either
random selection or maximization of the variational free energy $\mathcal{L}(q)$
over them. The latter procedure can be conducted by means of any appropriate
iterative maximization algorithm; here, we employ the popular L-BFGS
algorithm \cite{l-bfgs} for this purpose. Both random $\hat{\boldsymbol{x}}_{c}$
selection and estimation by means of variational free energy optimization,
using the L-BFGS algorithm, shall be evaluated in the experimental
section of our paper.

\section*{Acknowledgment}

The authors would like to thank Dr. David B. Dunson for the enlightening
discussion regarding the correct way to implement the MCMC sampler
for the KSBP.

\bibliographystyle{IEEEtran}

\begin{thebibliography}{10}
\providecommand{\url}[1]{#1}
\csname url@samestyle\endcsname
\providecommand{\newblock}{\relax}
\providecommand{\bibinfo}[2]{#2}
\providecommand{\BIBentrySTDinterwordspacing}{\spaceskip=0pt\relax}
\providecommand{\BIBentryALTinterwordstretchfactor}{4}
\providecommand{\BIBentryALTinterwordspacing}{\spaceskip=\fontdimen2\font plus
\BIBentryALTinterwordstretchfactor\fontdimen3\font minus
  \fontdimen4\font\relax}
\providecommand{\BIBforeignlanguage}[2]{{%
\expandafter\ifx\csname l@#1\endcsname\relax
\typeout{** WARNING: IEEEtran.bst: No hyphenation pattern has been}%
\typeout{** loaded for the language `#1'. Using the pattern for}%
\typeout{** the default language instead.}%
\else
\language=\csname l@#1\endcsname
\fi
#2}}
\providecommand{\BIBdecl}{\relax}
\BIBdecl

\bibitem{dpm2}
S.~Walker, P.~Damien, P.~Laud, and A.~Smith, ``Bayesian nonparametric inference
  for random distributions and related functions,'' \emph{J. Roy. Statist. Soc.
  B}, vol.~61, no.~3, pp. 485--527, 1999.

\bibitem{dpm3}
R.~Neal, ``Markov chain sampling methods for {Dirichlet} process mixture
  models,'' \emph{J. Comput. Graph. Statist.}, vol.~9, pp. 249--265, 2000.

\bibitem{dpm4}
P.~Muller and F.~Quintana, ``Nonparametric {Bayesian} data analysis,''
  \emph{Statist. Sci.}, vol.~19, no.~1, pp. 95--110, 2004.

\bibitem{DPM}
C.~Antoniak, ``Mixtures of {Dirichlet} processes with applications to
  {Bayesian} nonparametric problems.'' \emph{The Annals of Statistics}, vol.~2,
  no.~6, pp. 1152--1174, 1974.

\bibitem{bj}
D.~Blei and M.~Jordan, ``Variational methods for the {Dirichlet} process,'' in
  \emph{21st Int. Conf. Machine Learning}, New York, NY, USA, July 2004, pp.
  12--19.

\bibitem{py}
J.~Pitman and M.~Yor, ``The two-parameter {Poisson-Dirichlet} distribution
  derived from a stable subordinator,'' in \emph{Annals of Probability},
  vol.~25, 1997, pp. 855--900.

\bibitem{goldwater}
S.~Goldwater, T.~Griffiths, and M.~Johnson, ``Interpolating between types and
  tokens by estimating power-law generators,'' in \emph{Advances in Neural
  Information Processing Systems}, vol.~18, 2006.

\bibitem{sudderth}
E.~B. Sudderth and M.~I. Jordan, ``Shared segmentation of natural scenes using
  dependent pitman-yor processes,'' in \emph{Advances in Neural Information
  Processing Systems}, 2008, pp. 1585--1592.

\bibitem{ksbp}
D.~B. Dunson and J.-H. Park, ``Kernel stick-breaking processes,''
  \emph{Biometrika}, vol.~95, pp. 307--323, 2007.

\bibitem{hdp}
Y.~W. Teh, M.~I. Jordan, M.~J. Beal, and D.~M. Blei, ``Sharing clusters among
  related groups: Hierarchical {Dirichlet} processes,'' in \emph{Advances in
  Neural Information Processing Systems (NIPS)}, 2005, pp. 1385--1392.

\bibitem{ferguson-dp}
T.~Ferguson, ``A {Bayesian} analysis of some nonparametric problems,''
  \emph{The Annals of Statistics}, vol.~1, pp. 209--230, 1973.

\bibitem{polya-urn}
D.~Blackwell and J.~MacQueen, ``Ferguson distributions via {P\'olya} urn
  schemes,'' \emph{The Annals of Statistics}, vol.~1, no.~2, pp. 353--355,
  1973.

\bibitem{teh}
Y.~W. Teh, ``A hierarchical {Bayesian} language model based on {Pitman-Yor}
  processes,'' in \emph{Proc. Association for Computational Linguistics}, 2006,
  pp. 985--992.

\bibitem{stick-break}
J.~Sethuraman, ``A constructive definition of the {Dirichlet} prior,''
  \emph{Statistica Sinica}, vol.~2, pp. 639--650, 1994.

\bibitem{vbdpm}
D.~M. Blei and M.~I. Jordan, ``Variational inference for {Dirichlet} process
  mixtures,'' \emph{Bayesian Analysis}, vol.~1, no.~1, pp. 121--144, 2006.

\bibitem{dpm-hmm}
Y.~Qi, J.~W. Paisley, and L.~Carin, ``Music analysis using hidden {Markov}
  mixture models,'' \emph{IEEE Transactions on Signal Processing}, vol.~55,
  no.~11, pp. 5209--5224, 2007.

\bibitem{BishopBook}
C.~M. Bishop, \emph{Pattern Recognition and Machine Learning}.\hskip 1em plus
  0.5em minus 0.4em\relax New York: Springer, 2006.

\bibitem{l-bfgs}
D.~Liu and J.~Nocedal, ``On the limited memory method for large scale
  optimization,'' \emph{Mathematical Programming B}, vol.~45, no.~3, pp.
  503--528, 1989.

\bibitem{caruana}
R.~Caruana, ``Multitask learning,'' \emph{Machine Learning}, vol.~28, pp.
  41--75, 1997.

\bibitem{baxter}
J.~L. i.~r. Baxter, ``Learning internal representations,'' in \emph{COLT:
  Proceedings of the workshop on computational learning theory}, 1995.

\bibitem{evgeniou}
T.~Evgeniou, C.~Micchelli, and M.~Pontil, ``Learning multiple tasks with kernel
  methods,'' \emph{Journal of Machine Learning Research}, vol.~6, pp. 615--637,
  2005.

\bibitem{lawrence}
N.~Lawrence and J.~Platt, ``Learning to learn with the informative vector
  machine,'' in \emph{In Proceedings of the 21st International Conference on
  Machine Learning}, 2004.

\bibitem{yu}
K.~Yu, A.~Schwaighofer, V.~Tresp, W.-Y. Ma, and H.~Zhang, ``Collaborative
  ensemble learning: Combining collaborative and content-based information
  filtering via hierarchical bayes,'' in \emph{In Proceedings of the 19th
  conference on uncertainty in artificial intelligence}, 2003.

\bibitem{yu2}
K.~Yu, A.~Schwaighofer, and V.~Tresp, ``Learning gaussian processes from
  multiple tasks,'' in \emph{In Proceedings of the 22nd international
  conference on machine learning}, 2005.

\bibitem{berkley-images}
D.~Martin, C.~Fowlkes, D.~Tal, and J.~Malik, ``A database of human segmented
  natural images and its application to evaluating segmentation algorithms and
  measuring ecological statistics,'' in \emph{Proc. 8th Int'l Conf. Computer
  Vision}, Vancouver, Canada, July 2001, pp. 416--423.

\bibitem{hksbp}
Q.~An, C.~Wang, I.~Shterev, E.~Wang, L.~Carin, and D.~B. Dunson,
  ``{Hierarchical kernel stick-breaking process for multi-task image
  analysis},'' \emph{in Proceedings of the 25th international conference on
  Machine learning - ICML '08}, pp. 17--24, 2008.

\bibitem{PR}
R.~Unnikrishnan, C.~Pantofaru, and M.~Hebert, ``A measure for objective
  evaluation of image segmentation algorithms,'' in \emph{Proc. IEEE Conf.
  Computer Vision and Pattern Recognition}, San Diego, CA, USA, June 2005, pp.
  34--41.

\bibitem{normalized}
------, ``Toward objective evaluation of image segmentation algorithms,''
  \emph{IEEE Transactions on Pattern Analysis and Machine Intelligence},
  vol.~29, no.~6, pp. 929--944, 2007.

\bibitem{superpix}
G.~Mori, ``Guiding model search using segmentation.'' in \emph{Proc. 10th IEEE
  Int. Conf. on Computer Vision (ICCV)}, 2005.

\bibitem{filterbanks}
M.~Varma and A.~Zisserman, ``Classifying images of materials: Achieving
  view-point and illumination independence.'' in \emph{Proc. 7th IEEE European
  Conf. on Computer Vision (ECCV)}, 2002.

\bibitem{japanese}
M.~Kudo, J.~Toyama, and M.~Shimbo, ``Multidimensional curve classification
  using passing-through regions,'' \emph{Pattern Recognition Letters}, vol.~20,
  no. 11-13, pp. 1103--1111, 1999.

\bibitem{uci}
\BIBentryALTinterwordspacing
A.~Asuncion and D.~Newman, ``{UCI} machine learning repository,'' 2007.
  [Online]. Available:
  \url{http://www.ics.uci.edu/$\sim$mlearn/{MLR}epository.html}
\BIBentrySTDinterwordspacing

\end{thebibliography}


\end{document}